\documentclass[10pt,twocolumn,letterpaper]{article}

\usepackage[pagenumbers]{wacv} 

\usepackage{graphicx}
\usepackage{amsmath}
\usepackage{amssymb}
\usepackage{booktabs}
\usepackage{tikz}
\usepackage{comment}
\usepackage{amsmath,amssymb} 
\usepackage{textcomp}
\usepackage{color}
\usepackage{bm}
\usepackage{algorithmic}
\usepackage{setspace}
\usepackage{array}
\usepackage{multirow}
\usepackage{colortbl}
\usepackage{wrapfig}
\usepackage[accsupp]{axessibility}
\newcommand{\1}{\mbox{1}\hspace{-0.25em}\mbox{l}}

\usepackage[pagebackref,breaklinks,colorlinks]{hyperref}

\usepackage[capitalize]{cleveref}
\crefname{section}{Sec.}{Secs.}
\Crefname{section}{Section}{Sections}
\Crefname{table}{Table}{Tables}
\crefname{table}{Tab.}{Tabs.}

\def\wacvPaperID{832} 
\def\confName{WACV}
\def\confYear{2025}

\begin{document}

\title{ComFace: Facial Representation Learning with Synthetic Data\\ for Comparing Faces}

\author{Yusuke Akamatsu$^{1,\dagger}$, Terumi Umematsu$^{1}$, Hitoshi Imaoka$^{1}$, Shizuko Gomi$^{2}$, Hideo Tsurushima$^{2}$\\
$^{1}$NEC Corporation, Japan \qquad $^{2}$University of Tsukuba, Japan\\
{\tt\small $^{\dagger}$yusuke-akamatsu@nec.com}
}
\maketitle

\begin{abstract}
Daily monitoring of intra-personal facial changes associated with health and emotional conditions has great potential to be useful for medical, healthcare, and emotion recognition fields.
However, the approach for capturing intra-personal facial changes is relatively unexplored due to the difficulty of collecting temporally changing face images.
In this paper, we propose a facial representation learning method using synthetic images for comparing faces, called {\it ComFace}, which is designed to capture intra-personal facial changes.
For effective representation learning, ComFace aims to acquire two feature representations, {\it i.e.}, inter-personal facial differences and intra-personal facial changes.
The key point of our method is the use of synthetic face images to overcome the limitations of collecting real intra-personal face images.
Facial representations learned by ComFace are transferred to three extensive downstream tasks for comparing faces: estimating facial expression changes, weight changes, and age changes from two face images of the same individual.
Our ComFace, trained using only synthetic data, achieves comparable to or better transfer performance than general pre-training and state-of-the-art representation learning methods trained using real images.
\end{abstract}

\section{Introduction}
\label{sec:intro}

Human faces contain a variety of information, including identities, health conditions, and emotions.
Face recognition has long been studied for personal identification~\cite{zhao2003face,wang2021deep}.
Besides face recognition, biological information such as age~\cite{zhang2019c3ae,deng2021pml}, facial expression~\cite{goodfellow2013challenges,li2020deep}, body weight~\cite{dantcheva2018show,akamatsu2022edema}, and body mass index (BMI)~\cite{dantcheva2018show,sidhpura2022face} have also been estimated from face images.
Estimation of biological information from faces has great potential to be applied to medical, healthcare, and emotion recognition fields.
In these fields, it is important to capture daily intra-personal changes associated with health and emotional conditions~\cite{henly2011health,farhud2015impact,trampe2015emotions,zhang2018intra}.
Specifically, monitoring daily weight~\cite{akamatsu2022edema} and facial expression changes~\cite{kondo2021siamese} from the face helps to understand a person's health and emotional state.

To estimate biological information, most previous methods have performed classification or regression analysis from a single face image.
For example, a face image is classified into a facial expression class~\cite{li2020deep}, or BMI is estimated from a face image~\cite{sidhpura2022face}.
However, these methods do not focus on capturing subtle intra-personal changes in health and emotions.
Since it is useful to understand the subtle changes, capturing temporal changes in the face within an individual is important.
One of the reasons estimating intra-personal changes from face images has remained relatively unexplored is that it is difficult to collect a large number of temporally changing face images.

Many existing methods for face analysis have used deep neural networks (DNNs) with supervised learning.
To overcome the limitations of training data annotated with supervised labels, the mainstream approach utilizes DNNs pre-trained on large-scale data ({\it e.g.}, ImageNet~\cite{deng2009imagenet} and VGGFace2~\cite{cao2018vggface2}) and then performs transfer learning with annotated face data.
Also, facial representation learning (FRL) has been recently attracting attention~\cite{bulat2022pre,zheng2022general,liu2023pose,xue2023unsupervised}, where a large amount of unlabeled face images~\cite{bulat2022pre,liu2023pose,xue2023unsupervised} and face image-text pairs~\cite{zheng2022general} are used to obtain prior knowledge about facial representation and improve the performance of downstream face tasks.
Nevertheless, existing FRL methods mainly learn representations of facial differences between individuals and neglect those of facial changes within an individual.
Furthermore, visual representation learning using synthetic images has recently emerged in the computer vision domain~\cite{tian2023stablerep,tian2023learning}, surpassing the performance of representations learned using real images.
Even in the face domain, representation learning using synthetic images for temporally changing faces, which are very scarce in real images, has the potential to boost the estimation performance of intra-personal facial changes.

\begin{figure*}[t]
\begin{center}
\includegraphics[scale=0.4]{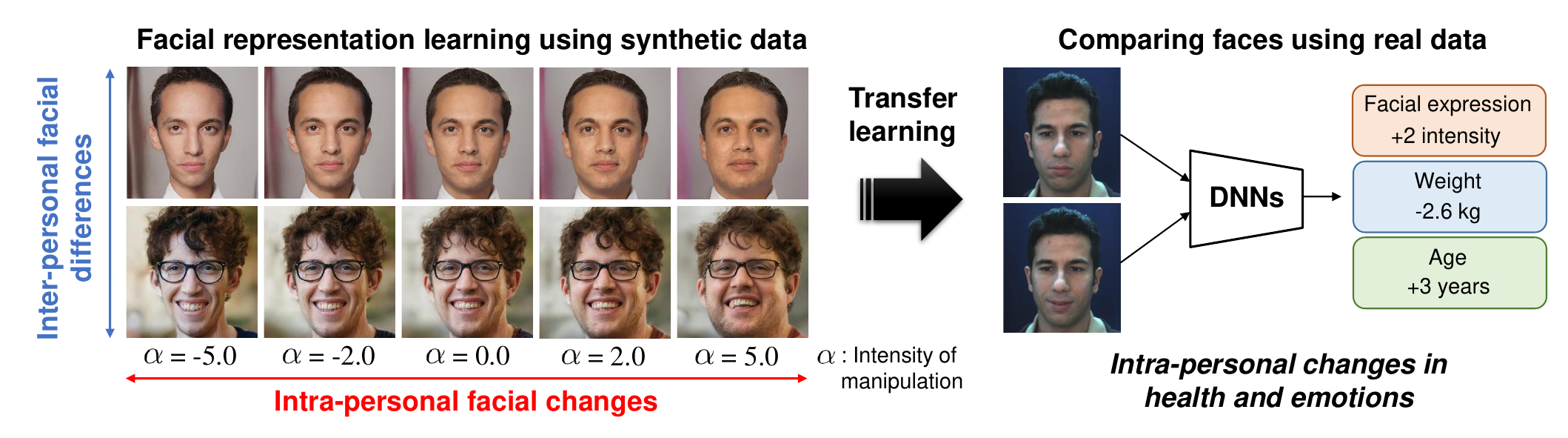}
\end{center}
\vspace{-20pt}
\caption{Overview of ComFace framework. ComFace performs facial representation learning using synthetic data relating to inter-personal facial differences and intra-personal facial changes. Then, facial representations are transferred to downstream tasks for comparing faces in order to capture intra-personal facial changes.} \vspace{-5pt}
\label{fig:intro}
\end{figure*}

In this paper, we propose an FRL method using synthetic data for comparing faces, called ComFace, which is designed to capture intra-personal facial changes.
Figure~\ref{fig:intro} is an overview of ComFace framework.
ComFace aims to learn two feature representations using synthetic data: inter-personal facial differences and intra-personal facial changes.
This makes it feasible to acquire facial representations relating not only to facial differences between individuals, but also to subtle facial changes within an individual.
To address the limitation in the number of real intra-personal face images, ComFace utilizes synthetic face images generated by StyleGANs~\cite{karras2019style,karras2020training,karras2020analyzing,karras2021alias}.
We generate synthetic face images for a large number of individuals and then generate intra-personal face images that vary according to many facial attributes.
Benefiting from synthetic data, ComFace can use an unlimited number of individuals and intra-personal face images, overcoming the problem of existing real image datasets with a small number of temporally changing face images.

The first FRL in ComFace is self-supervised contrastive learning~\cite{chen2020simple} to identify facial differences between individuals.
With self-supervised learning, ComFace learns feature representations of inter-personal facial differences without supervised labels.
In the second FRL, ComFace learns feature representations of intra-personal facial changes.
Specifically, how much the intra-personal face has changed is learned from two face images of the same individual.
Furthermore, curriculum learning~\cite{bengio2009curriculum} is introduced to gradually increase the difficulty level of estimating intra-personal facial changes during training, resulting in successful representation learning related to subtle facial changes.

We transfer DNNs pre-trained by ComFace to three downstream tasks for comparing faces: estimating facial expression changes, weight changes, and age changes from two face images of the same individual.
In these downstream tasks, we aim to estimate the direction and degree of facial expression change ({\it e.g.}, +2 and -3 intensity), weight change ({\it e.g.}, +2.6 kg and -1.3 kg), and age change ({\it e.g.}, +10 and -5 years) from two face images.
For facial expression change, we estimate the change in intensity of facial expressions for action unit 6 (AU6) and AU12.
For weight change, we estimate the weight change associated with the degree of facial edema in dialysis patients, as shown in a recent study~\cite{akamatsu2022edema}.
These downstream tasks consist of various periods of temporal changes in the face: short ({\it i.e.}, facial expression change), medium ({\it i.e.}, weight change), and long ({\it i.e.}, age change) periods.

The main contributions of this paper are summarized as follows:
\vspace{-5pt}
\begin{itemize}
\item We propose ComFace, the first FRL method using synthetic face images for comparing faces. ComFace learns feature representations regarding intra-personal facial changes as well as inter-personal facial differences.
\vspace{-5pt}
\item Facial representations learned by ComFace are transferred to three extensive downstream tasks for comparing faces. ComFace achieves comparable to or superior transfer performance to general pre-training and state-of-the-art (SoTA) representation learning methods trained using real images.
\vspace{-5pt}
\item Our approach of comparing two face images within an individual generalizes well to new patients and environmental conditions not used for training data.
Our weight change estimation model, trained without patient-specific data, outperforms the previous method~\cite{akamatsu2022edema} that estimates weight from a single face image, trained with patient-specific data.
\end{itemize}

\section{Related Work}
{\bf Face Analysis:} Various biological information has been estimated from a single face image~\cite{zhang2019c3ae,akamatsu2022edema,sidhpura2022face,mavadati2013disfa}.
Akamatsu {\it et al.}~\cite{akamatsu2022edema} focused on edema, a symptom of kidney disease, and estimated weight from the degree of facial edema in dialysis patients.
Weight reflects the fluid volume in a dialysis patient's body, so it is helpful to be able to easily monitor daily weight from a face image.
The weakness of their method is that it requires training data from a patient who uses the system in order to construct a patient-specific model.
This is because the method~\cite{akamatsu2022edema} estimates weight from a single face image and does not generalize well to new patients not used for training data.
In contrast, our approach of comparing two intra-personal face images generalizes well to new patients and does not require patient-specific data.
Our motivation to capture daily intra-personal facial changes is similar to Ref.~\cite{kondo2021siamese}, which estimates changes in the degree of smiling from two face images.
While the method~\cite{kondo2021siamese} performs transfer learning using a pre-trained model for face recognition, we use a pre-trained model specialized to capture intra-personal facial changes on the basis of synthetic images and achieve better transfer performance.
\\
\vspace{-10pt}
\\
{\bf Facial Representation Learning:}
Transfer learning in face analysis tasks commonly depends on pre-training using ImageNet~\cite{deng2021pml,akamatsu2022edema,ng2015deep,lin2020feasibility} and large face recognition datasets~\cite{sidhpura2022face,thinh2021emotion,parkin2019recognizing,knyazev2017convolutional,ranjan2017all}.
Recently, facial representation learning (FRL) has been studied as a pre-training method for face analysis tasks~\cite{bulat2022pre,zheng2022general,liu2023pose}.
Bulat {\it et al.}~\cite{bulat2022pre} investigated pre-training strategies and datasets for several face analysis tasks. 
Zheng {\it et al.}~\cite{zheng2022general} proposed a weakly-supervised method called FaRL using face image-text pairs.
Liu {\it et al.}~\cite{liu2023pose} proposed a pose-disentangled contrastive learning (PCL) method for general self-supervised facial representation.
Different from the FRL methods described above, our ComFace focuses on FRL regarding intra-personal facial changes.
Inter- and intra-personal influences on human interactions have been studied previously~\cite{kim2023hiint,woo2023amii}.
These methods simply handle inter- and intra-personal multimodal signals ($e.g.,$ video~\cite{kim2023hiint}, audio~\cite{kim2023hiint,woo2023amii}, facial gestures~\cite{woo2023amii}) to consider human-human interactions for specific applications such as modeling affect dynamics and adapted behavior synthesis.
Our work differs from previous work~\cite{kim2023hiint,woo2023amii} in that our work is a comprehensive representation learning study for downstream tasks that capture intra-personal facial changes.
\\
\vspace{-10pt}
\\
{\bf Synthetic Data:} 
Synthetic data for human analysis has been widely explored~\cite{joshi2024synthetic}, {\it e.g.}, face recognition~\cite{kortylewski2018training,kortylewski2019analyzing,trigueros2021generating,zhai2021demodalizing,qiu2021synface}, crowd counting~\cite{wang2019learning,wang2021pixel}, and fingerprint recognition~\cite{irtem2019impact,engelsma2022printsgan}.
In particular, with the recent success of generative adversarial networks (GANs)~\cite{goodfellow2014generative}, the quality of face synthesis has improved rapidly.
Qiu {\it et al.}~\cite{qiu2021synface} proposed a face recognition method called SynFace using synthetic face images generated by DiscoFaceGAN~\cite{deng2020disentangled}.
They explored the performance gap between face recognition models trained with synthetic and real face images and then designed SynFace to suppress the domain gap between synthetic and real face images.
Most recently, representation learning using synthetic data has emerged~\cite{tian2023stablerep,tian2023learning,di2024pros}.
Ref.~\cite{di2024pros} is the first study to consider introducing synthetic images into FRL and compares the transfer performance with that of real images.
Ref.~\cite{di2024pros} aims to learn general facial representation and differs from our FRL that focuses on intra-personal facial changes.
Although synthetic data have been introduced into the face domain, their performance in Refs.~\cite{qiu2021synface,di2024pros} is still lower than when using real images.
Our motivation differs from that of Refs.~\cite{qiu2021synface,di2024pros}, {\it i.e.}, while Refs.~\cite{qiu2021synface,di2024pros} explores the potential of synthetic images instead of real images, we rather leverage synthetic images since we have very few real images for temporally changing faces.
We make the most of the advantages of synthetic images over real images, resulting in our FRL using synthetic images achieving better transfer performance than other methods using real images.

\vspace{-3pt}
\section{Method}
\vspace{-5pt}

\subsection{Synthetic Face Images}
\vspace{-5pt}
\label{subsec:synthetic}
Our method utilizes synthetic face images generated by StyleGANs~\cite{karras2019style,karras2020training,karras2020analyzing,karras2021alias} for representation learning.
StyleGAN is one of the most popular generative models for face synthesis, demonstrating impressive performance in image generation, inversion, and manipulation~\cite{alaluf2022third}.
StyleGAN generators synthesize exceedingly realistic images and enable editing~\cite{shen2020interpreting,collins2020editing,abdal2021styleflow,alaluf2021only,ling2021editgan,patashnik2021styleclip}.
InterFaceGAN~\cite{shen2020interpreting} is a major framework for face editing.
It semantically edits faces by interpreting latent semantics learned by StyleGAN.
Specifically, InterFaceGAN can easily edit latent code $\bm{w}$ in high dimensional space $\mathcal{W}$ to manipulate the attributes ({\it e.g.}, weight, age, and gender) of a synthesized image as $\bm{w}_{edit} = \bm{w} + \alpha \cdot \bm{n}_{m}$,
where $\bm{w}_{edit}$ is the edited latent code, $\bm{n}_{m}$ is a normal vector to manipulate for attribute $m$, and $\alpha$ is a parameter that controls the intensity of the manipulation.
It will make the synthesis look more positive for each attribute when $\alpha > 0$ ({\it e.g.}, get fat when the attribute $m$ is weight), and $\alpha < 0$ will make the synthesis look more negative ({\it e.g.}, slim down when the attribute $m$ is weight).
As attributes to manipulate synthetic faces, we use weight~\cite{pinnimty2020transforming}, age~\cite{shen2020interpreting}, smile~\cite{shen2020interpreting}, and the 40 attributes included in the CelebA dataset~\cite{liu2015faceattributes}.
Examples of synthetic images manipulated when the attribute $m$ is weight are shown on the left side of Fig.~\ref{fig:intro}.
We synthesize a new person's face by randomly calculating the latent code $\bm{w}$, and generate images of intra-personal face changes by setting multiple $\alpha$.

\begin{figure*}[t]
\begin{center}
\includegraphics[scale=0.38]{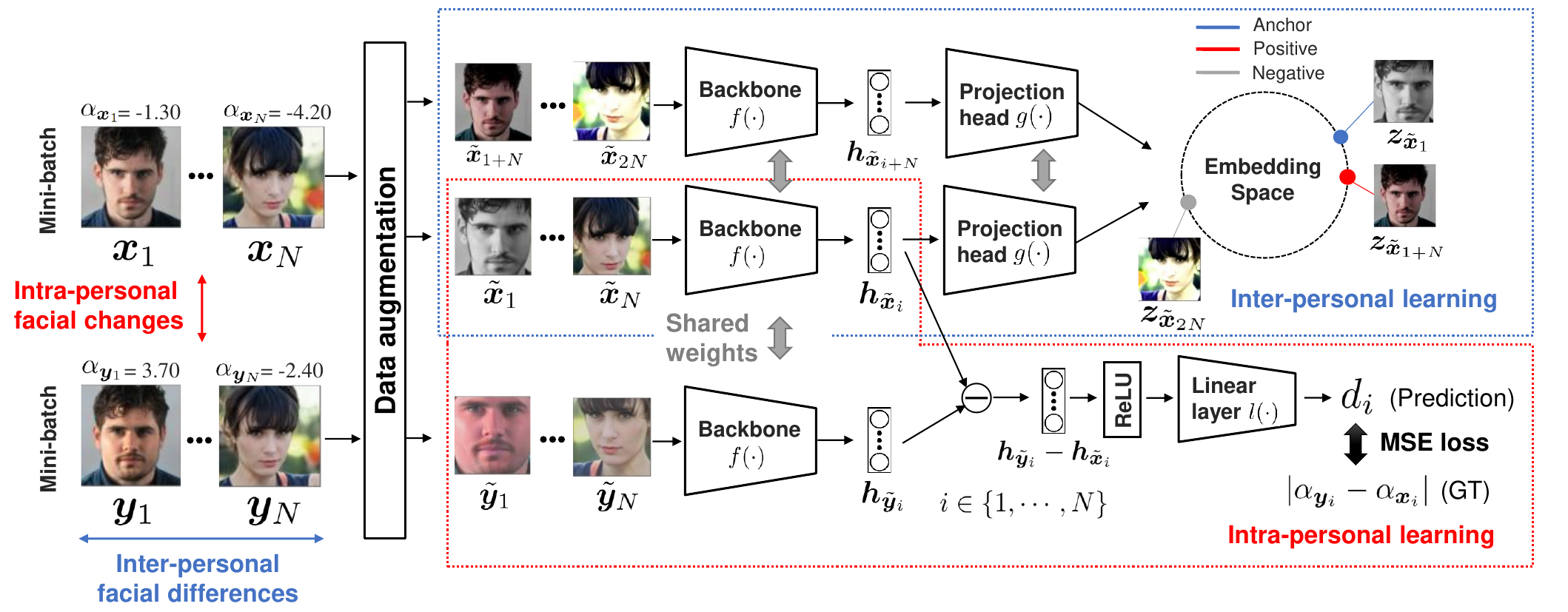}
\end{center}
\vspace{-15pt}
\caption{Training scheme of ComFace. Learning strategy consists of two components, {\it i.e.}, inter-personal learning and intra-personal learning. Inter-personal learning acquires feature representations of facial differences between individuals. Intra-personal learning acquires feature representations of facial changes within individuals.} \vspace{-5pt}
\label{fig:method}
\end{figure*}

\subsection{FRL with Synthetic Face Images}
\vspace{-5pt}
\label{subsec:frl}
Our ComFace learns two feature representations, {\it i.e.}, inter-personal facial differences and intra-personal facial changes.
Since the inter-personal represents broad differences in faces and the intra-personal represents subtle changes in faces, both of these two facial representations are essential to downstream tasks for comparing faces.
Figure~\ref{fig:method} shows the training scheme of ComFace.
The goal of the training is representation learning of the backbone $f(\cdot)$ using synthetic face images.
As the backbone $f(\cdot)$, arbitrary DNNs such as ResNet~\cite{he2016deep} and Vision Transformer (ViT)~\cite{dosovitskiy2020image} are used.
During the training, suppose that $\bm{x}_i$ is the $i$-th synthetic image in the mini-batch ($i=\{1,\cdots,N\}$, $N$ is the mini-batch size) and the parameter $\alpha$ corresponding to $\bm{x}_i$ is denoted as $\alpha_{\bm{x}_i}$ (see Section~\ref{subsec:synthetic}).
Also, $\bm{y}_i$ is a synthetic image edited from the face of the same person in $\bm{x}_i$, and the parameter $\alpha$ corresponding to $\bm{y}_i$ is denoted as $\alpha_{\bm{y}_i}$.
Since $\bm{x}_i$ and $\bm{y}_i$ are synthetic images of the same person whose face has been edited, the distance between intra-personal face changes is represented by $|\alpha_{\bm{y}_i}-\alpha_{\bm{x}_i}|$.
FRL in ComFace consists of the following inter-personal learning and intra-personal learning.
\\
{\bf Inter-personal Learning:} We use contrastive learning to acquire feature representations of facial differences between individuals.
Contrastive learning is a kind of self-supervised learning that performs representation learning without supervised labels~\cite{chen2020simple,he2020momentum}, attracting positive pairs ({\it e.g.}, the same images with different data augmentation) and pulling negative pairs ({\it e.g.}, different images) away.
ComFace performs contrastive learning based on SimCLR~\cite{chen2020simple}.
First, the synthetic image $\bm{x}_i$ is converted into two images $\tilde{\bm{x}}_{i}$ and $\tilde{\bm{x}}_{i+N}$ by data augmentation consisting of horizontal flip, color jitter, grayscale conversion, and random crop.
Since $\tilde{\bm{x}}_{i}$ and $\tilde{\bm{x}}_{i+N}$ are derived from the same image, they are referred to as a positive pair.
Then, $\tilde{\bm{x}}_{i}$ is input to $f(\cdot)$, which outputs the vector $\bm{h}_{\tilde{\bm{x}}_{i}}=f(\tilde{\bm{x}}_{i})$.
A projection head $g(\cdot)$ consisting of a two-layer multi-layer perceptron (MLP) follows after $f(\cdot)$, and we input $\bm{h}_{\tilde{\bm{x}}_{i}}$ and output vector $\bm{z}_{\tilde{\bm{x}}_{i}}=g(\bm{h}_{\tilde{\bm{x}}_{i}})$.
A mini-batch consists of $N$ samples, and data augmentation creates pairs, resulting in a total of $2N$ samples.
In contrastive learning, one positive pair $[\bm{z}_{\tilde{\bm{x}}_{i}},\bm{z}_{\tilde{\bm{x}}_{i+N}}]$ and the other $2(N-1)$ negative pairs are consisted, and the positive pairs are attracted and the negative ones are pulled away from each other.
Specifically, we perform contrastive learning based on the following InfoNCE loss~\cite{van2018representation}:
\vspace{-3pt}
\begin{flalign}
    \mathcal{L}_{inter} = - \frac{1}{2N} \sum_{j=1}^{2N} \log \frac{\exp(\bm{z}_{\tilde{\bm{x}}_{j}}\cdot \bm{z}_{\tilde{\bm{x}}_{j}}^p /\tau)}{\sum_{k=1}^{2N} \1_{k \neq j} \exp(\bm{z}_{\tilde{\bm{x}}_{j}}\cdot \bm{z}_{\tilde{\bm{x}}_{k}} /\tau)},
\end{flalign}
where $\bm{z}_{\tilde{\bm{x}}_{j}}^p$ is a vector that forms a positive pair with $\bm{z}_{\tilde{\bm{x}}_{j}}$ ({\it i.e.}, $\bm{z}_{\tilde{\bm{x}}_{i}}$ and $\bm{z}_{\tilde{\bm{x}}_{i+N}}$), $\1_{k\neq j} \in \{0,1\}$ is a function that returns 1 when $k\neq j$ and 0 when $k=j$, and $\tau$ is a temperature parameter.
With the above optimization, feature vectors of the same face are brought closer and those of different faces are kept apart.
\\
\vspace{-10pt}
\\
{\bf Intra-personal Learning:} We acquire feature representations of intra-personal facial changes.
The synthetic image $\bm{y}_i$ is converted into the image $\tilde{\bm{y}}_{i}$ by data augmentation.
Then, $\tilde{\bm{y}}_{i}$ is input to $f(\cdot)$, which outputs the vector $\bm{h}_{\tilde{\bm{y}}_{i}}=f(\tilde{\bm{y}}_{i})$.
We next calculate the difference vector $\bm{h}_{\tilde{\bm{y}}_{i}}-\bm{h}_{\tilde{\bm{x}}_{i}}$ between $\bm{h}_{\tilde{\bm{x}}_{i}}$ and $\bm{h}_{\tilde{\bm{y}}_{i}}$ to capture intra-personal facial changes.
Furthermore, the difference vector is input to ReLU and a linear layer $l(\cdot)$, which outputs the distance of intra-personal facial changes $d_i$.
ComFace learns how much the intra-personal face has changed on the basis of the mean squared error (MSE) loss: $\mathcal{L}_{intra} = \frac{1}{N} \sum_{i=1}^{N} (d_i -|\alpha_{\bm{y}_i}-\alpha_{\bm{x}_i}|)^2$,
where $|\alpha_{\bm{y}_i}-\alpha_{\bm{x}_i}|$ is the ground truth (GT) for the distance of intra-personal facial changes.
Since the direction of facial change depends on the attribute, we use $|\alpha_{\bm{y}_i}-\alpha_{\bm{x}_i}|$ instead of $\alpha_{\bm{y}_i}-\alpha_{\bm{x}_i}$ as the intra-personal GT.

As described above, ComFace acquires two feature representations, inter-personal facial differences and intra-personal facial changes, using the sum of those loss functions: $\mathcal{L} = \mathcal{L}_{inter} + \mathcal{L}_{intra}$.

\subsection{Curriculum Learning of Intra-personal Facial Changes}
In intra-personal learning, the difficulty level of learning tasks depends on $\alpha$.
Specifically, when $\alpha=0.0$ and $5.0$, there are significant differences in the facial changes, so it is easy to identify them (see the left side of Fig.~\ref{fig:intro}).
On the other hand, when $\alpha=0.0$ and $2.0$, the facial changes are small, so it is difficult to identify them (see the left side of Fig.~\ref{fig:intro}).
ComFace provides effective representation learning by increasing the difficulty level of learning tasks on the basis of curriculum learning.
Curriculum learning~\cite{bengio2009curriculum} mimics the human learning behavior of starting with simple tasks and gradually learning more complex concepts.
ComFace gradually decreases the distance between intra-personal
facial changes $|\alpha_{\bm{y}_i}-\alpha_{\bm{x}_i}|$ during training to acquire more effective feature representations.
Specifically, suppose that $S$ is the range of the distance of facial changes $|\alpha_{\bm{y}_i}-\alpha_{\bm{x}_i}|$ when sampling the pairs of $\bm{x}_i$ and $\bm{y}_i$ used for training, and we gradually decrease $S$ according to the number of epochs as follows:
\begin{equation}
S = \frac{S_{max}}{S_{e}}, \quad S_{e} = 
    \begin{cases}
        {1 \ (e \leq e_1)}\\
        {t \ (e_{t-1} < e \leq e_t)},
    \end{cases}
\end{equation}
where $S_{max}$ is the maximum range of the distance of facial changes in the dataset, and $e$ is the number of epochs.
In Eq. (2), we perform curriculum learning by increasing the difficulty level of learning tasks with the progress of the training ({\it i.e.}, with the increase in the number of epochs).

\subsection{Transfer Learning toward Downstream Tasks}
We transfer DNNs pre-trained by ComFace to downstream tasks for comparing faces.
For downstream tasks, we utilize DNNs in intra-personal learning (see the red box in Fig.~\ref{fig:method}). 
Since the linear layer $l(\cdot)$ trained by intra-personal learning is useful for comparing faces, we transfer both backbone $f(\cdot)$ and linear layer $l(\cdot)$.
In transfer learning, we perform downstream tasks for comparing faces using two face images, $\bm{x}^{task}_i$ and $\bm{y}^{task}_i$ ($i=\{1,\cdots,N^{task}$\}, $N^{task}$ is the number of images), and corresponding labels $\alpha_{\bm{x}^{task}_i}$ and $\alpha_{\bm{y}^{task}_i}$ ({\it e.g.}, facial expression intensity and weight) to estimate intra-personal facial changes.
Let $d^{task}_i$ be the predicted facial change; the DNN model is transferred on the basis of the MSE loss: $\mathcal{L}_{task} = \frac{1}{N^{task}} \sum_{i=1}^{N^{task}} (d_i^{task} -(\alpha_{\bm{y}^{task}_i}-\alpha_{\bm{x}^{task}_i}))^2$.
In each downstream task, we estimate not only the distance but also the direction of facial changes ({\it e.g.}, increase/decrease of weight).
Therefore, unlike the distance of the facial change in the intra-personal GT (see Section~\ref{subsec:frl}), we use $\alpha_{\bm{y}^{task}_i}-\alpha_{\bm{x}^{task}_i}$ as GT for downstream tasks.

\vspace{-3pt}
\section{Experiments}
\vspace{-3pt}
\subsection{Setup for FRL}
{\bf Synthetic Face Images:} We utilize synthetic face images generated by StyleGAN~\cite{karras2019style} and StyleGAN3~\cite{karras2021alias}.
The reason for using two different StyleGANs is to employ the attributes for face manipulation provided in each StyleGAN.
In StyleGAN, we use the attributes that vary weight provided by Ref.~\cite{pinnimty2020transforming} and the attributes that vary age and smile provided by InterFaceGAN~\cite{shen2020interpreting}.
In StyleGAN3, we employ the 40 attributes included in the CelebA dataset~\cite{liu2015faceattributes} provided by Ref.~\cite{alaluf2022third}.
The 40 attributes contain various face-related factors such as Big Nose, Bags Under Eyes, and Pale Skin.
With these 43 attributes, we use synthetic face images that change according to a wide range of face attributes.
We generate synthetic face images with 250,000 identities using StyleGANs.
For each identity, synthetic images for each attribute with $\alpha=\{-5.0,-4.9,\cdots,0.0,\cdots,4.9,5.0\}$ are generated (see Section~\ref{subsec:synthetic}).
As a result, the total number of synthetic face images used for FRL is 35 million (M).
For details on the composition of synthetic images, please refer to Supplement A.1.
See Supplement C.1 for a quality assessment of the synthetic images and an evaluation of the transfer performance with respect to the quality.
\\
\vspace{-5pt}
\\
{\bf Details for FRL:} For FRL with synthetic images, we use 90\% of all identities as training data and 10\% as validation data.
As a backbone $f(\cdot)$, we use ResNet50~\cite{he2016deep}, which is commonly used for general pre-training~\cite{deng2009imagenet,cao2018vggface2} and representation learning~\cite{chen2020simple,he2020momentum,bulat2022pre}.
Since ResNet50 is better than ViT~\cite{dosovitskiy2020image} for ComFace (see Supplement C.2 for the evaluation), we employ ResNet50.
Synthetic images are resized to 224$\times$224 and the temperature parameter $\tau$ is set to 0.1.
Our model is trained from scratch with randomly initialized weights.
We run the training for 12 epochs with batch size 1024 on 32 NVIDIA A100 GPUs ($\sim$22 hours training).
As parameters for curriculum learning, we set $S_{max}=10, \; t \in \{2,3,4\}, \; e_1=3, \;e_2=6, \;e_3=9, \;e_4=12$.
For each epoch, we randomly sample $\bm{x}_i$ and $\bm{y}_i$ pairs according to the range $S$ in curriculum learning and construct mini-batches.
Adam~\cite{kingma2014adam} optimizer is used, and the learning rate is initialized as 4e-4 and halved in 10 epochs.
See Supplement B.1 for other setups.

\subsection{Setup for Downstream Tasks}
We use the following facial expression change, weight change, and age change datasets in our downstream tasks for comparing faces.
The three downstream tasks correspond to short ($\sim$1 day), medium (1 day $\sim$ 3 months), and long (1 year$\sim$) time periods of temporal changes in the face, respectively.
\\
{\bf Facial Expression Change Dataset:}
We use the public dataset DISFA~\cite{mavadati2013disfa,mavadati2012automatic}, which contains facial videos of 27 subjects (12 women and 15 men) while they watch a 4-minute video intended to elicit a range of facial expressions.
For each subject, 4845 video frames were recorded, and the action unit (AU) intensity was annotated for each frame with six levels from 0 (not present) to 5 (maximum intensity) for several AUs.
We use AU6 (Cheek Raiser) and AU12 (Lip Corner Puller), which contain frames with a high AU intensity, and we extract frames with an intensity from 1 to 5.
We estimate the intensity changes of AU6 and AU12 as facial expression changes.
See Supplement C.3 for an evaluation for other major AUs.
\\
{\bf Weight Change Dataset:}
We use the dataset collected in Ref.~\cite{akamatsu2022edema} (Edema-A) and our newly collected dataset (Edema-B).
These datasets contain face images and weight data obtained from dialysis patients before and after dialysis.
Edema-A and Edema-B were collected from different hospitals and different patients.
Dialysis removes fluid from the body, which generally results in 2 to 3 kg weight change, with edema appearing on the face before dialysis and edema being alleviated after dialysis~\cite{akamatsu2022edema}.
In Edema-A, the number of patients is 38, the total number of acquired data is 392 (including pre- and post-dialysis), and the total number of images is 39200 (using 100 images per data).
In Edema-B, the number of patients is 19, the total number of acquired data is 320, and the total number of images is 32000.
\\
{\bf Age Change Dataset:} We use the public dataset FG-NET~\cite{panis2016overview}, which contains 1002 face images from 82 subjects.
The age ranges from 0 to 69 years, and the number of images per subject is 12 on average.
\\
See Supplement A.2 for more information on the above datasets.
\\
\vspace{-5pt}
\\
{\bf Details for Downstream Tasks:}
For each dataset, we perform a four-fold cross-validation that splits subjects between the training and test data.
Hence, we evaluate the generalization performance for new subjects not included in the training data.
We use 10\% of the subjects in the training data as validation data.
For both training and testing, the estimation of intra-personal facial changes is performed using two randomly sampled face images within an individual ({\it i.e.}, approximately the same number of samples in the direction of increasing or decreasing changes).
Note that the sampled pairs are identical across methods.
For transfer learning, we use two types of evaluation, linear evaluation ({\it i.e.}, backbone $f(\cdot)$ is frozen and linear layer $l(\cdot)$ is trained from scratch~\footnote{See Supplement C.4 for an evaluation in other settings.}) and fine-tuning ({\it i.e.}, all layers in $f(\cdot)$ and $l(\cdot)$ are trained from pre-trained weights~\footnote{See Supplement C.5 for an evaluation in other settings.}), as in the previous representation learning studies~\cite{chen2020simple,he2020momentum,bulat2022pre}.
See Supplement B.2 for other setups.

\subsection{Comparative Methods and Evaluation Metrics}
We are curious about the following questions: {\it Does ComFace successfully acquire facial representations that capture intra-personal facial changes? How is the transfer performance of representation learning using synthetic images against real images?}
To answer these questions, we compare ComFace with four types of methods: scratch (random initialization), general pre-training, visual representation learning, and facial representation learning.
We use a supervised method using ImageNet~\cite{deng2009imagenet} and a face recognition method using VGGFace2~\cite{cao2018vggface2} for general pre-training. We also use SimCLR~\cite{chen2020simple}, MoCo v2~\cite{he2020momentum,chen2020improved}, SwAV~\cite{caron2020unsupervised}, and Barlow Twins~\cite{zbontar2021barlow} for visual representation learning.
In the above methods, we use ResNet50, which is the same backbone as ours for fair comparison.
For facial representation learning, we use Bulat {\it et al.}~\cite{bulat2022pre}, FaRL~\cite{zheng2022general}, and PCL~\cite{liu2023pose}.
Note that all comparative methods use real images to learn feature representations.
In Supplement B.3, we summarize the training datasets, training scales, training sources, and backbones for all comparative and proposed methods.

The downstream tasks for comparing faces are regression analyses that estimate facial expression, weight, and age changes from two face images.
As evaluation metrics, we use the mean absolute error (MAE) and Pearson correlation coefficient (Corr.).
In weight change estimation, we also use accuracy (Acc.), which represents the prediction performance in the direction of weight gain or loss, since weight changes before and after dialysis.

\setlength{\tabcolsep}{4pt}
\begin{table}[t]
\caption{Results of estimating facial expression change for AU6 and AU12. Results are evaluated in linear evaluation and fine-tuning.}
\vspace{-17pt}
\label{table:au}
\begin{center}
\scalebox{0.62}{
\begin{tabular}{lccccccccc}  \hline \noalign{\smallskip}
 & \multicolumn{4}{c}{AU6} & \multicolumn{4}{c}{AU12}
\\ \noalign{\smallskip}
 & \multicolumn{2}{c}{Linear} & \multicolumn{2}{c}{Fine-tuning} & \multicolumn{2}{c}{Linear} & \multicolumn{2}{c}{Fine-tuning}
 \\ \noalign{\smallskip}
 \hline \noalign{\smallskip}
Method & MAE$\downarrow$ & Corr.$\uparrow$ & MAE$\downarrow$ & Corr.$\uparrow$ & MAE$\downarrow$ & Corr.$\uparrow$ & MAE$\downarrow$ & Corr.$\uparrow$   \\ \noalign{\smallskip} \hline \noalign{\smallskip}
Scratch & - & - & 0.745 & 0.277 & - & - & 0.975 & 0.326      \\
\noalign{\smallskip} \hline \noalign{\smallskip}
\textit{General Pre-training}: & & &   \\
ImageNet~\cite{deng2009imagenet} & 0.752 & 0.339 & 0.650 & 0.613 & 0.973 & 0.457 & 0.634 & 0.801      \\
VGGFace2~\cite{cao2018vggface2} & 0.730 & 0.461 & 0.660 & 0.578 & 0.796 & 0.670 & 0.644 & 0.795      \\
\noalign{\smallskip} \hline \noalign{\smallskip}
\multicolumn{5}{l}{\textit{Visual Representation Learning}:} & & & &   \\
SimCLR~\cite{chen2020simple} & 0.742 & 0.469 & 0.662 & 0.595 & 0.959 & 0.576 & 0.664 & 0.793  \\
MoCo v2~\cite{he2020momentum,chen2020improved} & 0.749 & 0.294 & 0.621 & 0.639 & 0.985 & 0.484 & 0.606 & 0.815  \\
SwAV~\cite{caron2020unsupervised} & 0.745 & 0.464 & 0.626 & 0.655 & 0.964 & 0.645 & 0.656 & 0.796  \\
Barlow Twins~\cite{zbontar2021barlow} & 0.745 & 0.472 & 0.662 & 0.607 & 0.958 & 0.698 & 0.659 & 0.793 \\
\noalign{\smallskip}
\hline \noalign{\smallskip}
\multicolumn{2}{l}{\textit{Facial Representation Learning}:} &   \\
Bulat {\it et al.}~\cite{bulat2022pre} & 0.745 & 0.412 & \bf{0.604} & \bf{0.669} & 0.964 & 0.567 & 0.599 & 0.829    \\ 
FaRL~\cite{zheng2022general} & 0.727 & 0.476 & 0.645 & 0.627 & 0.809 & 0.748 & 0.617 & 0.820      \\ 
PCL~\cite{liu2023pose} & 0.735 & 0.415 & 0.684 & 0.552 & 0.914 & 0.635 & 0.636 & 0.800    \\ 
\bf{ComFace (Ours)} & \bf{0.639} & \bf{0.648} & 0.629 & 0.663 & \bf{0.663} & \bf{0.786} & \bf{0.598} & \bf{0.831}   \\ 
\noalign{\smallskip} \hline \noalign{\smallskip}
\end{tabular}
} \vspace{-20pt}
\end{center}
\end{table}
\setlength{\tabcolsep}{1.4pt}

\subsection{Main Results}
{\bf Facial Expression Change:} Table~\ref{table:au} shows the results of estimating facial expression change in AU6 and AU12.
First, in linear evaluation~\footnote{To evaluate the learned representations, the backbone is frozen and the linear layer is trained from scratch. For a fair comparison, this setting is the same for all methods.}, ComFace outperforms all other methods on both AUs by a large margin.
This result indicates the advantage of ComFace, which focuses on intra-personal facial changes, over other methods that do not focus on such changes.
Thus, we find that ComFace successfully acquires representations that capture intra-personal facial changes by using synthetic face images.
Second, in fine-tuning, ComFace is slightly worse for AU6 and slightly better for AU12 than Bulat {\it et al.}~\cite{bulat2022pre}.
Our model, trained using only synthetic images, has comparable transfer performance to the SoTA FRL method trained using real images, suggesting the potential of representation learning using synthetic images.
Furthermore, ComFace performs better than general pre-training and visual representation learning methods, so our synthetic image-based model outperforms previous baseline methods.
\\
\vspace{-5pt}
\\
{\bf Weight Change:} In facial expression change estimation, fine-tuning is superior to linear evaluation for ComFace.
Since our ultimate goal is achieving high performance, we perform an evaluation in fine-tuning for the rest of our experiments.
Table~\ref{table:weight} shows the results of weight change estimation for an Edema-A, Edema-B, and Edema-A$\rightarrow$B cross-dataset evaluation.
In the cross-dataset evaluation, models trained on Edema-A are tested directly on Edema-B.
This is done to evaluate the robustness of the models against differences in lighting/environmental conditions and patient groups across hospitals.
From table~\ref{table:weight}, we confirm that ComFace outperforms all other methods in most metrics.
Specifically, our model has an accuracy improvement of 3.8\%, 2.9\%, and 4.3\% over the best model for general pre-training, visual representation learning, and FRL for Edema-A and 4.4\%, 3.5\%, and 1.7\% for Edema-B, respectively.
Furthermore, ComFace has better performance than the other methods in the cross-dataset evaluation.
The performance of Edema-A$\rightarrow$B in ComFace is close to that of Edema-B, indicating its high robustness.
These results suggest that the transfer performance of our representation learning using synthetic images is better than that of other methods using real images.
We also confirm that ComFace outperforms a visual representation learning method using synthetic images~\cite{tian2023stablerep}. See Supplement C.6 for details.

\setlength{\tabcolsep}{4pt}
\begin{table}[t]
\caption{Results of weight change estimation in Edema-A, Edema-B, and Edema-A$\rightarrow$B cross-dataset evaluation. Results are evaluated in fine-tuning.}
\vspace{-15pt}
\label{table:weight}
\begin{center}
\scalebox{0.59}{
\begin{tabular}{lccccccccc}  \hline \noalign{\smallskip}
& \multicolumn{3}{c}{Edema-A} & \multicolumn{3}{c}{Edema-B} & \multicolumn{3}{c}{Edema-A$\rightarrow$B}
\\ \noalign{\smallskip}
\cline{2-10}
\noalign{\smallskip}
Method &  MAE$\downarrow$ & Corr.$\uparrow$ & Acc.$\uparrow$ & MAE$\downarrow$ & Corr.$\uparrow$ & Acc.$\uparrow$ & MAE$\downarrow$ & Corr.$\uparrow$ & Acc.$\uparrow$   \\ \noalign{\smallskip} \hline
\noalign{\smallskip}
Scratch &  1.768 & 0.416 & 68.5 & 1.682 & 0.671 & 83.0 & 2.139 & 0.508 & 72.8 \\
\noalign{\smallskip}
\hline \noalign{\smallskip}
\textit{General Pre-training}: & & & & & & & & &  \\
ImageNet~\cite{deng2009imagenet} &  1.486 & 0.655 & 83.6 & 1.665 & 0.703 & 91.9 & 1.728 & 0.787 & 91.7    \\
VGGFace2~\cite{cao2018vggface2} &  1.482 & 0.695 & 84.8 & 1.593 & 0.778 & 90.2 & 1.732 & 0.785 & 91.5  \\ \noalign{\smallskip}
\hline \noalign{\smallskip}
\multicolumn{5}{l}{\textit{Visual Representation Learning}:} & & & &   \\
SimCLR~\cite{chen2020simple} & 1.535 & 0.671 & 82.2 & 1.504 & 0.788 & 92.6 & 1.917 & 0.735 & 88.4  \\
MoCo v2~\cite{he2020momentum,chen2020improved} &  1.451 & 0.702 & 84.8 & 1.488 & 0.767 & 91.0 & 1.759 & 0.775 & 90.9 \\
SwAV~\cite{caron2020unsupervised} & 1.483 & 0.718 & 85.7 & \bf{1.448} & 0.788 & 92.8 & 1.771 & 0.803 & 91.9  \\
Barlow Twins~\cite{zbontar2021barlow} & 1.552 & 0.634 & 80.4 & 1.575 & 0.753 & 90.6 & 1.867 & 0.741 & 86.7  \\
\noalign{\smallskip}
\hline \noalign{\smallskip}
\multicolumn{2}{l}{\textit{Facial Representation Learning}:} & & & & & & & &   \\
Bulat {\it et al.}~\cite{bulat2022pre} &  1.462 & 0.697 & 84.3 & 1.479 & 0.791 & 94.6 & 1.714 & 0.801 & 91.1   \\ 
FaRL~\cite{zheng2022general} &  1.517 & 0.654 & 82.8 & 1.454 & 0.773 & 93.5 & 1.862 & 0.718 &  88.9 \\ 
PCL~\cite{liu2023pose} &  1.484 & 0.662 & 82.0 & 1.510 & 0.748 & 89.2 & 1.840 & 0.715 & 83.9  \\
\bf{ComFace (Ours)} &  {\bf 1.394} & {\bf 0.750} & {\bf 88.6} & 1.523 & {\bf 0.801} & {\bf 96.3} & {\bf 1.668} & {\bf 0.819} & {\bf 93.8}  \\ \noalign{\smallskip} \hline
\end{tabular}
} 
\end{center}
\end{table}
\setlength{\tabcolsep}{1.4pt}


\setlength{\tabcolsep}{4pt}
\begin{table}[t]
\vspace{-10pt}
\caption{Results of estimating weight change with patient-specific model and our patient-generic model. Patient-specific model is transferred on 1 to 3 days of per-patient data.}
\vspace{-15pt}
\label{table:sota-weight}
\begin{center}
\scalebox{0.75}{
\begin{tabular}{lccc}  \hline \noalign{\smallskip}
& \multicolumn{3}{c}{Edema-A}
\\ \noalign{\smallskip}
\cline{2-4}
\noalign{\smallskip}
Method &  MAE$\downarrow$ & Corr.$\uparrow$ & Acc.$\uparrow$   \\ \noalign{\smallskip} \hline \noalign{\smallskip}
Patient-specific (1-day)~\cite{akamatsu2022edema} & 1.629 & 0.473 & 75.6 \\
Patient-specific (2-day)~\cite{akamatsu2022edema} & 1.483 & 0.617 & 78.7 \\
Patient-specific (3-day)~\cite{akamatsu2022edema} & \bf{1.245} & 0.717 & 85.6 \\
\bf{Patient-generic (Ours)} & 1.304 & \bf{0.720} & \bf{86.6} \\
\noalign{\smallskip} \hline
\end{tabular}
} \vspace{-25pt}
\end{center}
\end{table}
\setlength{\tabcolsep}{1.4pt}

\setlength{\tabcolsep}{4pt}
\begin{table}[t]
\caption{Results of estimating age change. Results are evaluated in fine-tuning.}
\vspace{-20pt}
\label{table:age}
\begin{center}
\scalebox{0.75}{
\begin{tabular}{lcc}  \hline \noalign{\smallskip} 
Method &  MAE$\downarrow$ & Corr.$\uparrow$    \\ \noalign{\smallskip} \hline \noalign{\smallskip}
Scratch & 8.980 & 0.514 \\
ImageNet~\cite{deng2009imagenet} & 7.863 & 0.614 \\
SwAV~\cite{caron2020unsupervised} & 6.368 & 0.780 \\
FaRL~\cite{zheng2022general} & 5.249 & 0.851  \\
\bf{ComFace (Ours)} & \bf{4.914} & \bf{0.870}   \\
\noalign{\smallskip} \hline
\end{tabular}
} \vspace{-15pt}
\end{center}
\end{table}
\setlength{\tabcolsep}{1.4pt}


Furthermore, we compare ComFace with the previous method that estimates weight from a single face image~\cite{akamatsu2022edema}.
The previous method performs pre-training on multiple patient data and then builds patient-specific models via transfer learning on per-patient data.
As in the original paper~\cite{akamatsu2022edema}, the patient-specific model uses 24 patients for pre-training and 15 patients for transfer learning and testing, and ComFace uses the same 15 patients for testing.
Since the performance of patient-specific models depends on the number of dialysis days on per-patient data for transfer learning, we evaluate the performance when using per-patient data from 1 to 3 days.
Table~\ref{table:sota-weight} shows the results of weight change estimation with the patient-specific model and our patient-generic model.
Our patient-generic model performs better than the patient-specific model transferred on 1 or 2 days data and is comparable to the patient-specific model on 3 days data.
Since our patient-generic model is built without patient-specific data ({\it i.e.}, 0 days), we confirm the advantage of our two image-based method over the previous single image-based method~\cite{akamatsu2022edema}.
The results suggest that our weight change estimation from two face images within an individual generalizes well to new patients.
\\
\vspace{-5pt}
\\
{\bf Age Change:}
Table~\ref{table:age} shows the results of estimating age change in fine-tuning.
The best models from the four types of comparative methods are listed (see Supplement C.7 for full versions).
The table indicates that ComFace has a higher transfer performance for age change than all other methods.
We find that our method can be successfully adapted to the task related to long time periods (1 year$\sim$) of temporal changes in the face.

\setlength{\tabcolsep}{4pt}
\begin{table}[t]
\caption{Ablation study on ComFace. In ``Learning'' column, ``Both'' denotes both inter- and intra-personal learning. Line indicated in gray is our final setting.}
\vspace{-15pt}
\label{table:ablation}
\begin{center}
\footnotesize
\scalebox{0.72}{
\begin{tabular}{ccccccccc}  \hline 
\noalign{\smallskip}
  & \multirow{2}{*}{Learning} & \multirow{2}{*}{Intra-personal GT}  & \multirow{2}{*}{Curriculum}   & AU6 & AU12 & Edema-A & Edema-B & Age   \\ 
  & & & &   Corr.$\uparrow$ & Corr.$\uparrow$ & Acc.$\uparrow$ & Acc.$\uparrow$ & Corr.$\uparrow$   \\ \noalign{\smallskip} \hline \noalign{\smallskip}
 (a) & Inter- &  $|\alpha_{\bm{y}_i}-\alpha_{\bm{x}_i}|$  & $\checkmark$ &  0.593 & 0.829 & 79.5 & 89.0 & 0.782  \\
 (b) & Intra- &  $|\alpha_{\bm{y}_i}-\alpha_{\bm{x}_i}|$  & $\checkmark$ &  0.660 & 0.833 & 87.0 & 93.1 & 0.861 \\
 \rowcolor[rgb]{0.9, 0.9, 0.9}
 \bf{(c)} & Both &  $|\alpha_{\bm{y}_i}-\alpha_{\bm{x}_i}|$  & $\checkmark$ &  0.663 & 0.831 & 88.6 & 96.3 & 0.870 \\
 (d) & Both &  $\alpha_{\bm{y}_i}-\alpha_{\bm{x}_i}$ & $\checkmark$  &   0.662 & 0.818 & 82.8 & 91.0 & 0.830  \\
 (e) & Both &  $|\alpha_{\bm{y}_i}-\alpha_{\bm{x}_i}|$ & & 
 0.659 & 0.821 & 88.0 & 94.9 & 0.856 
 \\ \noalign{\smallskip} \hline
\end{tabular}
} \vspace{-20pt}
\end{center}
\end{table}
\setlength{\tabcolsep}{1.4pt}

\begin{figure*}[t]
\begin{center}
\includegraphics[scale=0.33]{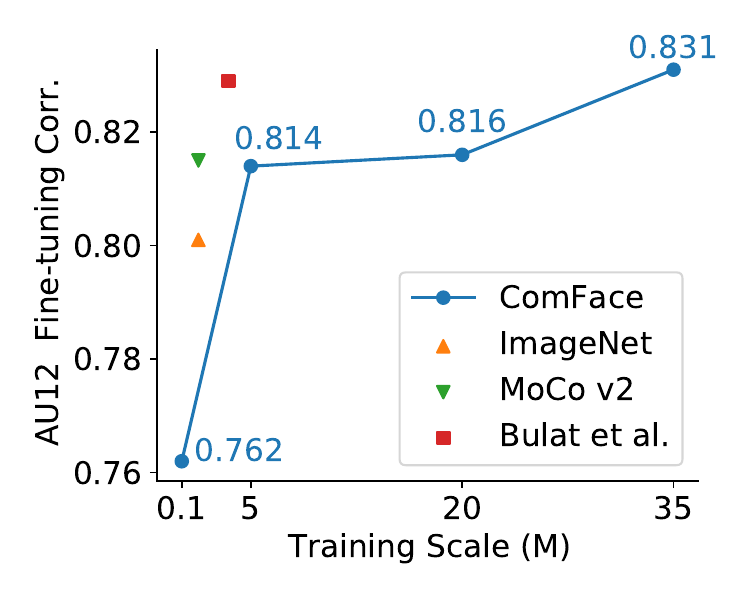}
\hspace{5pt}
\includegraphics[scale=0.33]{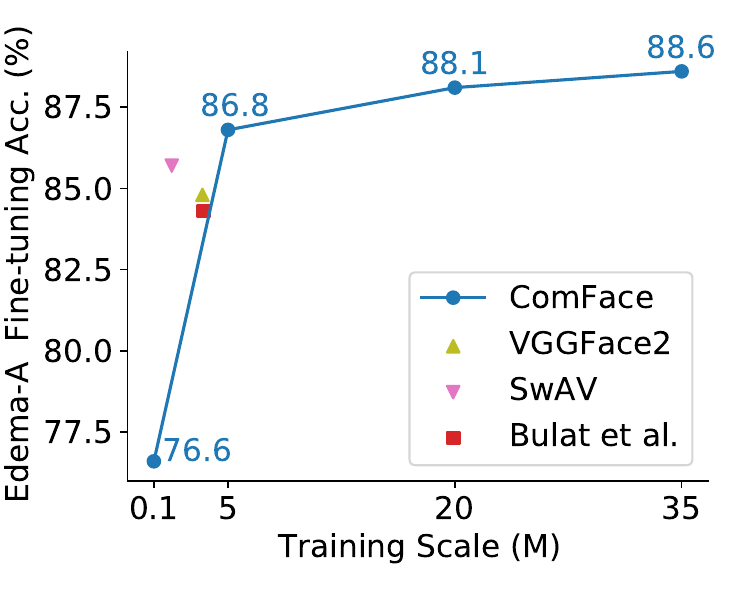}
\includegraphics[scale=0.33]{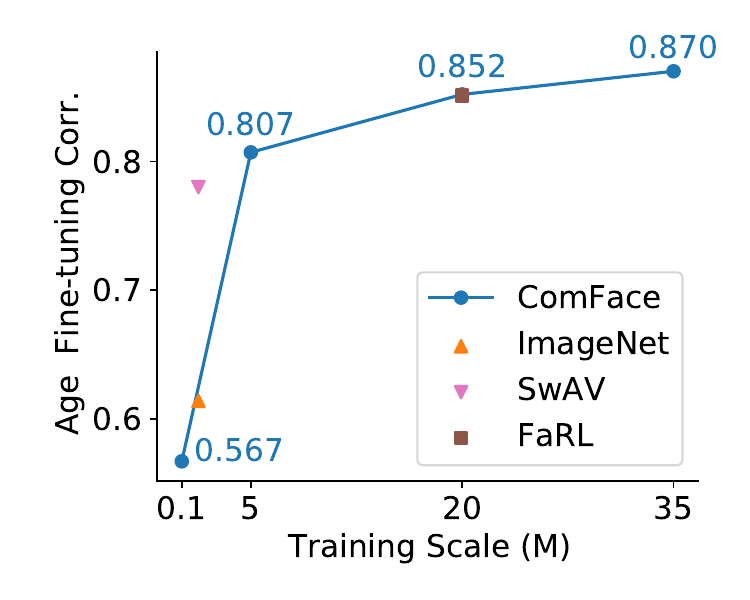}
\end{center}
\vspace{-20pt}
\caption{Transfer performance for facial expression change (AU12), weight change (Edema-A), and age change on different training scales. Facial expression change and age change are evaluated by correlation coefficient and weight change is evaluated by accuracy in fine-tuning. ComFace and the best models for general pre-training, visual representation learning, and FRL are compared.}
\label{fig:tr_scale}
\vspace{-10pt}
\end{figure*}

\begin{figure*}[t]
\begin{center}
\includegraphics[scale=0.33]{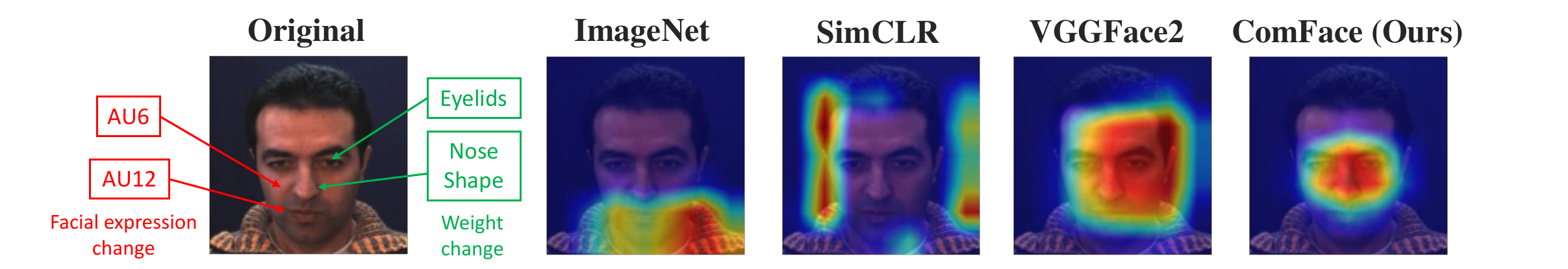}
\end{center}
\vspace{-15pt}
\caption{Saliency maps for a face image in four pre-trained backbones. Original images show positions of AU6 and AU12 and eyelid and nose shape where edema appears.}
\label{fig:cam}
\vspace{-10pt}
\end{figure*}

\subsection{Ablation Study}
\label{exp:ablation}
We ablate the components of ComFace to verify the effectiveness of each factor. 
In the ablation study, we evaluate transfer performance for three downstream tasks in fine-tuning.
Key observations are described as follows:
\\
{\bf Inter- and Intra-personal Learning:}
Table~\ref{table:ablation} (a,b,c) compares learning strategies.
It shows that intra-personal learning is more beneficial than inter-personal learning.
We find that representation learning that focuses on subtle facial changes is more essential than broad facial differences.
Furthermore, the employment of both inter- and intra-personal learning tends to improve the performance.
\\
{\bf Intra-personal GT:}
For FRL, we use $|\alpha_{\bm{y}_i}-\alpha_{\bm{x}_i}|$ as intra-personal GT (see Section~\ref{subsec:frl}) instead of $\alpha_{\bm{y}_i}-\alpha_{\bm{x}_i}$, which is used for transfer learning.
Table~\ref{table:ablation} (c,d) compares the two intra-personal GTs and shows that the use of $|\alpha_{\bm{y}_i}-\alpha_{\bm{x}_i}|$ is better than that of $\alpha_{\bm{y}_i}-\alpha_{\bm{x}_i}$.
We expect the use of $\alpha_{\bm{y}_i}-\alpha_{\bm{x}_i}$ harms FRL since the direction of facial change varies by attribute ({\it e.g.}, gender and age).
\\
{\bf Curriculum Learning:}
Table~\ref{table:ablation} (c,e) demonstrates the effectiveness of curriculum learning.
We can see that gradually increasing the difficulty level of identifying intra-personal facial changes via curriculum learning is reasonable and leads to effective representation learning.
\\
{\bf Training Scale:}
We investigate the performance of ComFace with respect to training scales of synthetic images.
We vary the number of identities for synthetic faces and set the following training scales: 0.1M, 5M, 20M, and 35M (our final setting).
Figure~\ref{fig:tr_scale} represents the transfer performance for three downstream tasks versus the training scales.
ComFace is compared with the best methods for general pre-training, visual representation learning, and FRL, respectively.
The performance of ComFace improves as the training scale increases.
ComFace outperforms the comparative method on the full training scale, showing the advantage of synthetic images not being limited in the number of images.



\subsection{Visualization}
We calculate saliency maps to provide the interpretability of models.
The saliency maps are obtained from the final block of the pre-trained backbones via Eigen-CAM~\cite{muhammad2020eigen}, which visualizes the principle components of learned representations without relying on class relevance scores.
Figure~\ref{fig:cam} illustrates the saliency maps of four models for a face image from DISFA~\cite{mavadati2013disfa,mavadati2012automatic}.
Models using ImageNet and SimCLR focus on regions other than the face ({\it e.g.}, clothing and hair), which is expected due to the fact that the models are trained on general images ({\it i.e.}, ImageNet).
The model using VGGFace2 focuses on the entire face since it is trained for face recognition.
Our ComFace focuses on the center of the face, which includes AU6, AU12, and eyelid and nose shapes that are affected by edema~\cite{akamatsu2022edema} (see the original image in Fig.~\ref{fig:cam}).
We believe that ComFace captures the center of the face, where intra-personal facial changes often appear, resulting in high transfer performance of downstream tasks for comparing faces.

\vspace{-3pt}
\section{Conclusion and Societal Impact}
This paper introduces facial representation learning (FRL) using synthetic images for comparing faces.
We obtain the following answers from the experimental results: (i) ComFace successfully acquires representations that capture intra-personal facial change. (ii) Our transfer performance in representation learning using synthetic images is comparable to or better than SoTA representation learning methods using real images. (iii) Our approach of comparing two face images generalizes well to new patients and environmental conditions.
In future work, we plan to explore more downstream tasks related to facial changes other than facial expression, weight, and age.
\\
{\bf Potential Negative Societal Impact:}
Machine learning models using face images may include biases such as ethnicity, age, and gender.
Leveraging synthetic images has the potential to reduce those biases by manipulating the generative model.
However, it is important to acknowledge that our method relies on generative models trained on images crawled from a large-scale website (Flicker).
The generative model inherits the biases of Flicker, and the synthetic images may contain social biases and errors.


\newpage 
{\small
\bibliographystyle{ieee_fullname}
\bibliography{main}

\begin{thebibliography}{10}\itemsep=-1pt

\bibitem{abdal2021styleflow}
R. Abdal, P. Zhu, N.~J. Mitra, and P. Wonka.
\newblock {StyleFlow}: Attribute-conditioned exploration of {StyleGAN}-generated images using conditional continuous normalizing flows.
\newblock {\em ACM Transactions on Graphics (ToG)}, 40(3):1--21, 2021.

\bibitem{akamatsu2022edema}
Y. Akamatsu, Y. Onishi, H. Imaoka, J. Kameyama, et~al.
\newblock Edema estimation from facial images taken before and after dialysis via contrastive multi-patient pre-training.
\newblock {\em IEEE Journal of Biomedical and Health Informatics}, 27(3):1419--1430, 2023.

\bibitem{alaluf2021only}
Y. Alaluf, O. Patashnik, and D. Cohen-Or.
\newblock Only a matter of style: Age transformation using a style-based regression model.
\newblock {\em ACM Transactions on Graphics (TOG)}, 40(4):1--12, 2021.

\bibitem{alaluf2022third}
Y. Alaluf, O. Patashnik, Z. Wu, A. Zamir, et~al.
\newblock Third time’s the charm? image and video editing with {StyleGAN3}.
\newblock In {\em Proc. European Conf. Computer Vision Workshops}, pages 204--220. Springer, 2022.

\bibitem{bengio2009curriculum}
Y. Bengio, J. Louradour, R. Collobert, and J. Weston.
\newblock Curriculum learning.
\newblock In {\em Proc. Int. Conf. Machine Learning (ICML)}, pages 41--48, 2009.

\bibitem{bulat2022pre}
A. Bulat, S. Cheng, J. Yang, A. Garbett, et~al.
\newblock Pre-training strategies and datasets for facial representation learning.
\newblock In {\em Proc. European Conf. Computer Vision (ECCV)}, pages 107--125. Springer, 2022.

\bibitem{cao2018vggface2}
Q. Cao, L. Shen, W. Xie, O.~M. Parkhi, et~al.
\newblock Vggface2: A dataset for recognising faces across pose and age.
\newblock In {\em Proc. Int. Conf. Automatic Face \& Gesture Recognition (FG)}, pages 67--74, 2018.

\bibitem{caron2020unsupervised}
M. Caron, I. Misra, J. Mairal, P. Goyal, et~al.
\newblock Unsupervised learning of visual features by contrasting cluster assignments.
\newblock In {\em Proc. Advances in Neural Information Processing Systems (NeurIPS)}, volume~33, pages 9912--9924, 2020.

\bibitem{chen2020simple}
T. Chen, S. Kornblith, M. Norouzi, and G. Hinton.
\newblock A simple framework for contrastive learning of visual representations.
\newblock In {\em Proc. Int. Conf. Machine Learning (ICML)}, pages 1597--1607, 2020.

\bibitem{chen2020improved}
X. Chen, H. Fan, R. Girshick, and K. He.
\newblock Improved baselines with momentum contrastive learning.
\newblock {\em arXiv preprint arXiv:2003.04297}, 2020.

\bibitem{collins2020editing}
E. Collins, R. Bala, B. Price, and S. Susstrunk.
\newblock Editing in style: Uncovering the local semantics of gans.
\newblock In {\em Proc. IEEE/CVF Conf. Computer Vision and Pattern Recognition (CVPR)}, pages 5771--5780, 2020.

\bibitem{dantcheva2018show}
A. Dantcheva, F. Bremond, and P. Bilinski.
\newblock Show me your face and {I} will tell you your height, weight and body mass index.
\newblock In {\em Proc. Int. Conf. Pattern Recognition (ICPR)}, pages 3555--3560. IEEE, 2018.

\bibitem{deng2009imagenet}
J. Deng, W. Dong, R. Socher, L. Li, et~al.
\newblock Imagenet: A large-scale hierarchical image database.
\newblock In {\em Proc. IEEE/CVF Conf. Computer Vision and Pattern Recognition (CVPR)}, pages 248--255, 2009.

\bibitem{deng2020disentangled}
Y. Deng, J. Yang, D. Chen, F. Wen, et~al.
\newblock Disentangled and controllable face image generation via 3d imitative-contrastive learning.
\newblock In {\em Proc. IEEE/CVF Conf. Computer Vision and Pattern Recognition (CVPR)}, pages 5154--5163, 2020.

\bibitem{deng2021pml}
Z. Deng, H. Liu, Y. Wang, C. Wang, et~al.
\newblock {PML}: Progressive margin loss for long-tailed age classification.
\newblock In {\em Proc. IEEE/CVF Conf. Computer Vision and Pattern Recognition (CVPR)}, pages 10503--10512, 2021.

\bibitem{di2024pros}
X. Di, Y. Zheng, X. Liu, and Y. Cheng.
\newblock Pros: Facial omni-representation learning via prototype-based self-distillation.
\newblock In {\em Proc. IEEE/CVF Winter Conf. Applications of Computer Vision (WACV)}, pages 6087--6098, 2024.

\bibitem{dosovitskiy2020image}
A. Dosovitskiy, L. Beyer, A. Kolesnikov, D. Weissenborn, et~al.
\newblock An image is worth 16x16 words: Transformers for image recognition at scale.
\newblock In {\em Proc. Int. Conf. Learning Representations (ICLR)}, 2020.

\bibitem{engelsma2022printsgan}
J.~J. Engelsma, S. Grosz, and A.~K. Jain.
\newblock Printsgan: Synthetic fingerprint generator.
\newblock {\em IEEE Transactions on Pattern Analysis and Machine Intelligence}, 45(5):6111--6124, 2022.

\bibitem{farhud2015impact}
D.~D. Farhud.
\newblock Impact of lifestyle on health.
\newblock {\em Iranian Journal of Public Health}, 44(11):1442, 2015.

\bibitem{goodfellow2014generative}
I. Goodfellow, J. Pouget-Abadie, M. Mirza, B. Xu, et~al.
\newblock Generative adversarial nets.
\newblock In {\em Proc. Advances in Neural Information Processing Systems (NeurIPS)}, volume~27, 2014.

\bibitem{goodfellow2013challenges}
I.~J. Goodfellow, D. Erhan, P.~L. Carrier, A. Courville, et~al.
\newblock Challenges in representation learning: A report on three machine learning contests.
\newblock In {\em Proc. Int. Conf. Neural Information Processing (ICONIP)}, pages 117--124, 2013.

\bibitem{he2020momentum}
K. He, H. Fan, Y. Wu, S. Xie, et~al.
\newblock Momentum contrast for unsupervised visual representation learning.
\newblock In {\em Proc. IEEE/CVF Conf. Computer Vision and Pattern Recognition (CVPR)}, pages 9729--9738, 2020.

\bibitem{he2016deep}
K. He, X. Zhang, S. Ren, and J. Sun.
\newblock Deep residual learning for image recognition.
\newblock In {\em Proc. IEEE/CVF Conf. Computer Vision and Pattern Recognition (CVPR)}, pages 770--778, 2016.

\bibitem{henly2011health}
S.~J. Henly, J.~F. Wyman, and M.~J. Findorff.
\newblock Health and illness over time: The trajectory perspective in nursing science.
\newblock {\em Nursing Research}, 60(3 Suppl):S5, 2011.

\bibitem{irtem2019impact}
P. Irtem, E. Irtem, and N. Erdo{\u{g}}mu{\c{s}}.
\newblock Impact of variations in synthetic training data on fingerprint classification.
\newblock In {\em Proc. Int. Conf. Biometrics Special Interest Group (BIOSIG)}, 2019.

\bibitem{joshi2024synthetic}
I. Joshi, M. Grimmer, C. Rathgeb, C. Busch, et~al.
\newblock Synthetic data in human analysis: A survey.
\newblock {\em IEEE Transactions on Pattern Analysis and Machine Intelligence}, 2024.

\bibitem{karras2020training}
T. Karras, M. Aittala, J. Hellsten, S. Laine, et~al.
\newblock Training generative adversarial networks with limited data.
\newblock In {\em Proc. Advances in Neural Information Processing Systems (NeurIPS)}, volume~33, pages 12104--12114, 2020.

\bibitem{karras2021alias}
T. Karras, M. Aittala, S. Laine, E. H{\"a}rk{\"o}nen, et~al.
\newblock Alias-free generative adversarial networks.
\newblock In {\em Proc. Advances in Neural Information Processing Systems (NeurIPS)}, volume~34, pages 852--863, 2021.

\bibitem{karras2019style}
T. Karras, S. Laine, and T. Aila.
\newblock A style-based generator architecture for generative adversarial networks.
\newblock In {\em Proc. IEEE/CVF Conf. Computer Vision and Pattern Recognition (CVPR)}, pages 4401--4410, 2019.

\bibitem{karras2020analyzing}
T. Karras, S. Laine, M. Aittala, J. Hellsten, et~al.
\newblock Analyzing and improving the image quality of {StyleGAN}.
\newblock In {\em Proc. IEEE/CVF Conf. Computer Vision and Pattern Recognition (CVPR)}, pages 8110--8119, 2020.

\bibitem{kim2023hiint}
Y. Kim, D.~W. Lee, P.~P. Liang, and S. Alghowinem.
\newblock {HIINT}: Historical, intra-and inter-personal dynamics modeling with cross-person memory transformer.
\newblock In {\em Proc. Int. Conf. Multimodal Interaction}, pages 314--325, 2023.

\bibitem{kingma2014adam}
D.~P. Kingma and J. Ba.
\newblock Adam: A method for stochastic optimization.
\newblock {\em arXiv preprint arXiv:1412.6980}, 2014.

\bibitem{knyazev2017convolutional}
B. Knyazev, R. Shvetsov, N. Efremova, and A. Kuharenko.
\newblock Convolutional neural networks pretrained on large face recognition datasets for emotion classification from video.
\newblock {\em arXiv preprint arXiv:1711.04598}, 2017.

\bibitem{kondo2021siamese}
K. Kondo, T. Nakamura, Y. Nakamura, and S. Satoh.
\newblock Siamese-structure deep neural network recognizing changes in facial expression according to the degree of smiling.
\newblock In {\em Proc. Int. Conf. Pattern Recognition (ICPR)}, pages 4605--4612, 2021.

\bibitem{kortylewski2019analyzing}
A. Kortylewski, B. Egger, A. Schneider, T. Gerig, et~al.
\newblock Analyzing and reducing the damage of dataset bias to face recognition with synthetic data.
\newblock In {\em Proc. IEEE/CVF Conf. Computer Vision and Pattern Recognition Workshops}, 2019.

\bibitem{kortylewski2018training}
A. Kortylewski, A. Schneider, T. Gerig, B. Egger, et~al.
\newblock Training deep face recognition systems with synthetic data.
\newblock {\em arXiv preprint arXiv:1802.05891}, 2018.

\bibitem{li2020deep}
S. Li and W. Deng.
\newblock Deep facial expression recognition: A survey.
\newblock {\em IEEE Transactions on Affective Computing}, 13(3):1195--1215, 2020.

\bibitem{lin2020feasibility}
S. Lin, Z. Li, B. Fu, S. Chen, et~al.
\newblock Feasibility of using deep learning to detect coronary artery disease based on facial photo.
\newblock {\em European Heart Journal}, 41(46):4400--4411, 2020.

\bibitem{ling2021editgan}
H. Ling, K. Kreis, D. Li, S.~W. Kim, et~al.
\newblock {EditGAN}: High-precision semantic image editing.
\newblock In {\em Proc. Advances in Neural Information Processing Systems (NeurIPS)}, volume~34, pages 16331--16345, 2021.

\bibitem{liu2023pose}
Y. Liu, W. Wang, Y. Zhan, S. Feng, et~al.
\newblock Pose-disentangled contrastive learning for self-supervised facial representation.
\newblock In {\em Proc. IEEE/CVF Conf. Computer Vision and Pattern Recognition (CVPR)}, pages 9717--9728, 2023.

\bibitem{liu2015faceattributes}
Z. Liu, P. Luo, X. Wang, and X. Tang.
\newblock Deep learning face attributes in the wild.
\newblock In {\em Proc. IEEE/CVF Int. Conf. Computer Vision (ICCV)}, 2015.

\bibitem{mavadati2012automatic}
S.~M. Mavadati, M.~H. Mahoor, K. Bartlett, and P. Trinh.
\newblock Automatic detection of non-posed facial action units.
\newblock In {\em Proc. Int. Conf. Image Processing (ICIP)}, pages 1817--1820, 2012.

\bibitem{mavadati2013disfa}
S.~M. Mavadati, M.~H. Mahoor, K. Bartlett, P. Trinh, et~al.
\newblock Disfa: A spontaneous facial action intensity database.
\newblock {\em IEEE Transactions on Affective Computing}, 4(2):151--160, 2013.

\bibitem{muhammad2020eigen}
M.~B. Muhammad and M. Yeasin.
\newblock {Eigen-CAM}: Class activation map using principal components.
\newblock In {\em Proc. Int. Joint Conf. Neural Networks (IJCNN)}, pages 1--7, 2020.

\bibitem{ng2015deep}
H. Ng, V.~D. Nguyen, V. Vonikakis, and S. Winkler.
\newblock Deep learning for emotion recognition on small datasets using transfer learning.
\newblock In {\em Proc. ACM on Int. Conf. Multimodal Interaction}, pages 443--449, 2015.

\bibitem{panis2016overview}
G. Panis, A. Lanitis, N. Tsapatsoulis, and T.~F. Cootes.
\newblock Overview of research on facial ageing using the {FG-NET} ageing database.
\newblock {\em IET Biometrics}, 5(2):37--46, 2016.

\bibitem{parkin2019recognizing}
A. Parkin and O. Grinchuk.
\newblock Recognizing multi-modal face spoofing with face recognition networks.
\newblock In {\em Proc. IEEE/CVF Conf. Computer Vision and Pattern Recognition Workshops}, 2019.

\bibitem{patashnik2021styleclip}
O. Patashnik, Z. Wu, E. Shechtman, D. Cohen-Or, et~al.
\newblock {StyleCLIP}: Text-driven manipulation of {StyleGAN} imagery.
\newblock In {\em Proc. IEEE/CVF Int. Conf. Computer Vision (ICCV)}, pages 2085--2094, 2021.

\bibitem{pinnimty2020transforming}
V. Pinnimty, M. Zhao, P. Achananuparp, and E. Lim.
\newblock Transforming facial weight of real images by editing latent space of {StyleGAN}.
\newblock {\em arXiv preprint arXiv:2011.02606}, 2020.

\bibitem{qiu2021synface}
H. Qiu, B. Yu, D. Gong, Z. Li, et~al.
\newblock {SynFace}: Face recognition with synthetic data.
\newblock In {\em Proc. IEEE/CVF Int. Conf. Computer Vision (ICCV)}, pages 10880--10890, 2021.

\bibitem{ranjan2017all}
R. Ranjan, S. Sankaranarayanan, C.~D. Castillo, and R. Chellappa.
\newblock An all-in-one convolutional neural network for face analysis.
\newblock In {\em Proc. Int. Conf. Automatic Face \& Gesture Recognition (FG)}, pages 17--24. IEEE, 2017.

\bibitem{shen2020interpreting}
Y. Shen, J. Gu, X. Tang, and B. Zhou.
\newblock Interpreting the latent space of {GANs} for semantic face editing.
\newblock In {\em Proc. IEEE/CVF Conf. Computer Vision and Pattern Recognition (CVPR)}, pages 9243--9252, 2020.

\bibitem{sidhpura2022face}
J. Sidhpura, R. Veerkhare, P. Shah, and S. Dholay.
\newblock {Face To BMI}: A deep learning based approach for computing bmi from face.
\newblock In {\em Proc. Int. Conf. Innovative Trends in Information Technology (ICITIIT)}, pages 1--6. IEEE, 2022.

\bibitem{thinh2021emotion}
P.~T.~D. Thinh, H.~M. Hung, H. Yang, S. Kim, et~al.
\newblock Emotion recognition with sequential multi-task learning technique.
\newblock In {\em Proc. IEEE/CVF Int. Conf. Computer Vision Workshops}, pages 3593--3596, 2021.

\bibitem{tian2023learning}
Y. Tian, L. Fan, K. Chen, D. Katabi, et~al.
\newblock Learning vision from models rivals learning vision from data.
\newblock {\em arXiv preprint arXiv:2312.17742}, 2023.

\bibitem{tian2023stablerep}
Y. Tian, L. Fan, P. Isola, H. Chang, and D. Krishnan.
\newblock {StableRep}: Synthetic images from text-to-image models make strong visual representation learners.
\newblock In {\em Proc. Advances in Neural Information Processing Systems (NeurIPS)}, 2023.

\bibitem{trampe2015emotions}
D. Trampe, J. Quoidbach, and M. Taquet.
\newblock Emotions in everyday life.
\newblock {\em PloS one}, 10(12):e0145450, 2015.

\bibitem{trigueros2021generating}
D. Trigueros, L. Meng, and M. Hartnett.
\newblock Generating photo-realistic training data to improve face recognition accuracy.
\newblock {\em Neural Networks}, 134:86--94, 2021.

\bibitem{van2018representation}
A. Van~den Oord, Y. Li, and O. Vinyals.
\newblock Representation learning with contrastive predictive coding.
\newblock {\em arXiv preprint arXiv:1807.03748}, 2018.

\bibitem{wang2021deep}
M. Wang and W. Deng.
\newblock Deep face recognition: A survey.
\newblock {\em Neurocomputing}, 429:215--244, 2021.

\bibitem{wang2019learning}
Q. Wang, J. Gao, W. Lin, and Y. Yuan.
\newblock Learning from synthetic data for crowd counting in the wild.
\newblock In {\em Proc. IEEE/CVF Conf. Computer Vision and Pattern Recognition (CVPR)}, pages 8198--8207, 2019.

\bibitem{wang2021pixel}
Q. Wang, J. Gao, W. Lin, and Y. Yuan.
\newblock Pixel-wise crowd understanding via synthetic data.
\newblock {\em International Journal of Computer Vision}, 129(1):225--245, 2021.

\bibitem{woo2023amii}
J. Woo, M. Fares, C. Pelachaud, and C. Achard.
\newblock {AMII}: Adaptive multimodal inter-personal and intra-personal model for adapted behavior synthesis.
\newblock {\em arXiv preprint arXiv:2305.11310}, 2023.

\bibitem{xue2023unsupervised}
F. Xue, Y. Sun, and Y. Yang.
\newblock Unsupervised facial expression representation learning with contrastive local warping.
\newblock {\em arXiv preprint arXiv:2303.09034}, 2023.

\bibitem{zbontar2021barlow}
J. Zbontar, L. Jing, I. Misra, Y. LeCun, et~al.
\newblock {Barlow Twins}: Self-supervised learning via redundancy reduction.
\newblock In {\em Proc. Int. Conf. Machine Learning (ICML)}, pages 12310--12320, 2021.

\bibitem{zhai2021demodalizing}
Z. Zhai, P. Yang, X. Zhang, M. Huang, et~al.
\newblock Demodalizing face recognition with synthetic samples.
\newblock In {\em Proc. AAAI Conf. Artificial Intelligence}, volume~35, pages 3278--3286, 2021.

\bibitem{zhang2019c3ae}
C. Zhang, S. Liu, X. Xu, and C. Zhu.
\newblock {C3AE}: Exploring the limits of compact model for age estimation.
\newblock In {\em Proc. IEEE/CVF Conf. Computer Vision and Pattern Recognition (CVPR)}, pages 12587--12596, 2019.

\bibitem{zhang2018intra}
S. Zhang, L. Baams, D. van~de Bongardt, and J.~S. Dubas.
\newblock Intra-and inter-individual differences in adolescent depressive mood: The role of relationships with parents and friends.
\newblock {\em Journal of Abnormal Child Psychology}, 46:811--824, 2018.

\bibitem{zhao2003face}
W. Zhao, R. Chellappa, P.~J. Phillips, and A. Rosenfeld.
\newblock Face recognition: A literature survey.
\newblock {\em ACM Computing Surveys (CSUR)}, 35(4):399--458, 2003.

\bibitem{zheng2022general}
Y. Zheng, H. Yang, T. Zhang, J. Bao, et~al.
\newblock General facial representation learning in a visual-linguistic manner.
\newblock In {\em Proc. IEEE/CVF Conf. Computer Vision and Pattern Recognition (CVPR)}, pages 18697--18709, 2022.

\end{thebibliography}
}

\end{document}